%% file: emnlp2021.tex
\pdfoutput=1
\documentclass[11pt]{article}

\usepackage[]{emnlp2021}
\usepackage{times}
\usepackage{latexsym}
\usepackage{graphicx}
\usepackage[ruled,vlined]{algorithm2e}
\usepackage{bbm}
\usepackage{mathtools}
\usepackage{amssymb, bm}
\usepackage{pifont}
\usepackage{float}
\usepackage{caption}
\usepackage{subcaption}
\usepackage{mathrsfs}
\usepackage{booktabs}
\usepackage{float}
\usepackage{tabulary}

\graphicspath{ {./sections/figs/} } 
\input{math_shortcuts}

\usepackage{times}
\usepackage{latexsym}

\usepackage[T1]{fontenc}
\usepackage[utf8]{inputenc}

\usepackage{microtype}

\title{Unsupervised Multiple Choices Question Answering: \\
Start Learning from Basic Knowledge}

\author{Chi-Liang Liu \quad  Hung-yi Lee \\ College of Electrical Engineering and Computer Science\\ National Taiwan University \\ {\tt \{liangtaiwan1230, tlkagkb93901106\}@gmail.com}\\}

\date{}

\begin{document}
\maketitle 

\begin{abstract}
In this paper, we study the possibility of unsupervised  Multiple Choices Question Answering (MCQA).
From very basic knowledge, the MCQA model knows that some choices have higher probabilities of being correct than others.
The information, though very noisy, guides the training of an MCQA model.
The proposed method is shown to outperform the baseline approaches on RACE and is even comparable with some supervised learning approaches on MC500.
\end{abstract}

\input{sections/intro}
\input{sections/unqa}

\input{sections/exp}

\input{sections/conclusion}

% \section*{Broader Impact}
% We explored the possibilities that generating pseudo labels from basic rules or external datasets. Furthermore, we show the  proposed  method significantly outperform the baseline approaches. We

\bibliography{emnlp2021}
\bibliographystyle{acl_natbib}

\clearpage
\appendix
\input{sections/appendix}

\end{document}

%% file: math_shortcuts.tex
\newcommand*{\Eb}{\mathbb{E}}

\newcommand*{\Lc}{\mathcal{L}}

\DeclarePairedDelimiterXPP{\KL}[2]{D_\text{KL}}\lbrack\rbrack{}{{#1} \delimsize\Vert {#2}}
\DeclarePairedDelimiterXPP{\Prob}[1]{\Pb}\lbrace\rbrace{}{#1}
\DeclarePairedDelimiterXPP{\Ev}[1]{\Eb}\lbrack\rbrack{}{#1}
\DeclarePairedDelimiterXPP{\Evr}[2]{\Eb_{#1}}\lbrack\rbrack{}{#2}

%% file: sections/intro.tex
\section{Introduction}

\textit{Question Answering} (QA) has been widely used for testing Reading Comprehension. Recently, numerous question answering datasets~\citep{weston2015towards, squad, squad2, yang2018hotpotqa, trischler2017newsqa, choi2018quac, joshi2017triviaqa, kwiatkowski2019natural, reddy2019coqa, mctest, lai2017race, khashabi-etal-2018-looking} have been proposed. These datasets can be divided into two major categories: \textit{Extractive Question Answering} (EQA) and \textit{Multiple Choices Question Answering} (MCQA).
In EQA, the answer has to be a span of the given reading passage, such as SQuAD~\citep{squad} and NewsQA~\citep{trischler2017newsqa}; while in MCQA, the answer is one of the given choices, such as MCTest~\citep{mctest} and RACE~\citep{lai2017race}.

Recently, large pretrained language models such as BERT~\citep{devlin2019bert} have exceeded human performance in some EQA benchmark corpora, for example, SQuAD~\citep{squad}. Compared to EQA, MCQA does not restrict the answer to be spans in context. This allowed MCQA can have more challenging questions than EQA, including but not limited to logical reasoning or summarization.
The performance gap between BERT and human performance is still significant. 
In this paper, we focus on MCQA.

A person who can read can deal with the MCQA task without further training, but this is not the case for a machine.
The BERT-based models cannot be directly applied to solve the MCQA task without seeing any MCQA examples.
Even for the models achieving human-level performance in EQA, they still need some MCQA examples with correct choices being labeled for fine-tuning. 
Although \citet{keskar2019unifying, raffel2020exploring} proposed the unified question answering model, they require unifying the multiple tasks to span extraction task. 

The semi-supervised MCQA model training approach has been proposed~\citep{chung-etal-2018-supervised}, in which an initial MCQA model is used to answer the unlabelled questions to generate pseudo labeled data. %ref to this one https://www.aclweb.org/anthology/N18-1143.pdf
Then pseudo labeled data is used to fine-tune the MCQA model to improve the performance.
However, the initial MCQA model still needs some labeled examples to train.

\begin{figure}[t!]
    \includegraphics[width=\columnwidth]{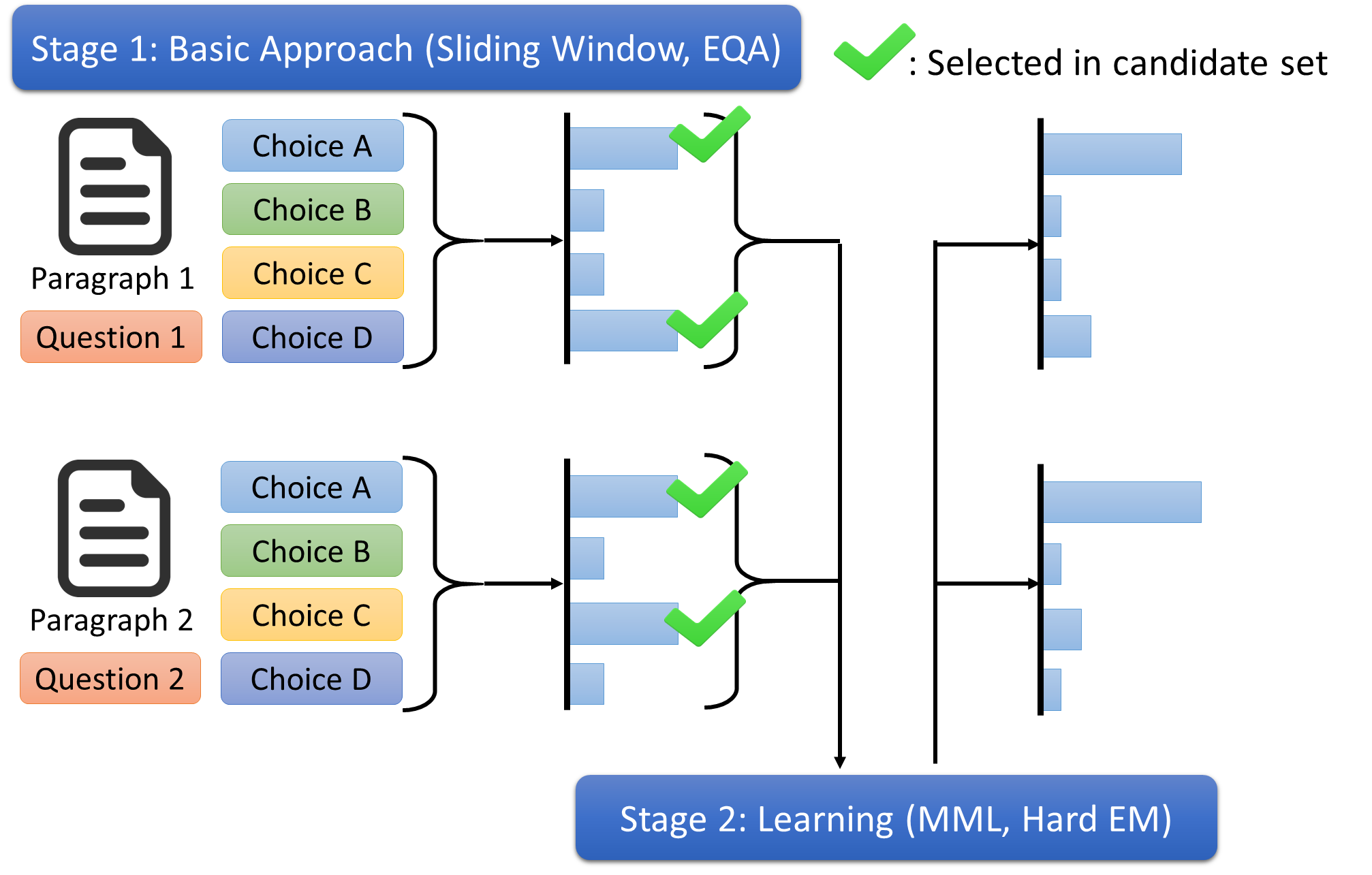}
    \caption{Overall training process.}
    \label{fig:pipline}
    \vspace{-0.5cm}
\end{figure}

In this paper, we study the possibility of unsupervised MCQA.
Instead of starting from an initial MCQA model~\citep{chung-etal-2018-supervised}, here, the machine starts with some prior knowledge. 
For example, a choice has a higher probability of being correct if the choice has word overlap with the document and question.
With the basic rule, the machine knows that some choices have higher probabilities of being correct than others, and some choices can be ruled out.
With these basic rules, an MCQA model can be trained without any labeled MCQA examples. 
With this approach, we got absolute gains of 4$\sim$9\% accuracy compared to the baseline methods on two MCQA benchmark corpora, RACE and MC500. 

%Related work: Hard EM, etc.

%% file: sections/unqa.tex
\section{Unsupervised MCQA}

We consider MCQA where we are given a question $q$, a passage $p$ and a set of choices $C=\{c_1, c_2,...c_n\}$, where $n$ is the number of choices, and machine needs to select an answer $a\in C$. 

We propose to address an unsupervised MCQA in a two-stage approach (Figure~\ref{fig:pipline}). 
First, we pick the candidate set $\mathcal{T}$ from choices by fundamental rule from human knowledge (sliding window) or a model trained without MCQA data (EQA model). 
Second, we train a model to pick the final answer from the candidates.

\subsection{Candidates Choosing}
The candidate selection approaches give a score to each choice which represents the likelihood of being correct.
We use two systems to calculate the scores, one using simple lexical features and another using a pre-trained EQA model. 
A choice is selected into candidate set $\mathcal{T}$ if the choice's score is higher than a threshold $t$, and is the top $k$ scores among all the choices $\{c_1, c_2,...c_n\}$ of a question $q$. 
In this way, each question has at most $k$ candidates in $\mathcal{T}$.
$k$ should be smaller than $n$ ($k<n$) to rule out some less likely choices.
A question will not have any choice in $\mathcal{T}$ if none of its choices pass the threshold $t$. %Should I explain more?
Both $t$ and $k$ are the hyperparameters. 
Note that our methods do \textbf{not} guarantee the answer must be in the candidate set.
The candidate sets are only used during training, and \textit{we do not need to choose candidates when testing}.

\paragraph{Sliding Window (SW)}
We follow the sliding window algorithm in \citet{mctest}, matching a bag of words constructed from the question and choices to the passage to compute the scores of choices. The algorithm's details are shown in Algorithm~\ref{alg:sw}.

\paragraph{EQA Matching}
In this setting, we use a pre-trained EQA model as our reference.
Given a passage and a question, the EQA model outputs an answer $A$, which is a text span from the passage.  
Then we use a string-matching algorithm to compute the similarity between $A$ and each candidate $c$, and the similarity serves as the score for each candidate.
Gestalt Pattern Matching~\citep{slidematch} algorithm is the string-matching algorithm used here.
The algorithm's details are shown in Appendix~\ref{sec:alg}.

\subsection{Learning Methods}
The candidates $\mathcal{T}$ selected in the last subsection are used as the ground truth to train an MCQA model.
Because the candidates are not always correct, and each question can have multiple choices selected in the candidate set, the typical supervised learning approaches cannot be directly applied here. 
Therefore, the following learning methods are explored to form our objective function $\mathcal{L}$ for training the MCQA model from the candidates.
\vspace{-0.1cm}
\paragraph{Highest-Only}
\[\Lc = -\log P\left(c_{max}\mid p; q\right),\vspace{-0.25cm}\]
where $c_{max}$ is the choice of a question $q$ in the candidate set  with the highest score.
The approach here has no difference from typical supervised learning, except that the ground truth is from the candidate selection approaches, not human labeling. 
\vspace{-0.1cm}
\paragraph{Maximum Marginal Likelihood (MML)}
\[\Lc = -\log \sum_{c_i\in \mathcal{T}}P\left(c_i\mid p; q\right)\vspace{-0.1cm}\]
In this objective, all the choices in the candidate set are considered correct.
The learning target of the MCQA model is to maximize the probabilities that all the choices in the candidate set are labeled as correct.
If there are more correct choices than the incorrect ones in the candidate set, the impact of the wrong choices in the candidate set can be mitigated.
\vspace{-0.1cm}
\paragraph{Hard-EM} Proposed by ~\citet{min2019discrete}, this can be viewed as a variant of MML,
\[\Lc = -\log \max_{c_i\in\mathcal{T}}P\left(c_i\mid p; q\right)\vspace{-0.1cm}\]
The underlying assumption of this objective can be understood as follows.
For a question $q$, several choices are selected in the candidate set.
Although we don't know which one is correct, we assume one of them is correct.  
Therefore, we want the MCQA model to learn to maximize the probability of one of the choices for a question.

%% file: sections/exp.tex
\section{Experiments Setup}
\input{sections/tables/results}
\input{sections/tables/candidates}

\input{sections/tables/chosen}

To evaluate the proposed method's effectiveness compared to supervised learning and other approaches that do not require training data, we experiment on two MCQA tasks, RACE and MCTest(MC500).

\subsection{Datasets}
\paragraph{RACE} \citet{race} introduced the RACE dataset, collected from the English exams for middle and high school Chinese students. RACE consists of near 28000 passages and nearly 100000 questions. Specifically, the dataset can be split into two parts: RACE-M, collected from English examinations designed for middle school students; and RACE-H, collected from English examinations designed for high students. RACE-H is more difficult than RACE-M; the length of the passages and the vocabulary size in the RACE-H are much larger than that of the RACE-M.
\paragraph{MC500} \citet{mctest} present MCTest which requires machines to answer multiple-choice reading comprehension questions about fictional stories. MCTest has two variants: MC160, which contains 160 stories, and MC500, which contains 500 stories. Moreover, MC500 can be subdivided into MC500-One and MC500-Multi. MC500-One refers to the questions that can be answered with one sentence. MC500-Multi refers to the questions that need evidence in multiple sentences to answer.

The length of each story is approximately 150 to 300 words, and the topic of a story is a wide range. %including vacations, animals, school, cars, eating,gardening, fairy tales, spaceships, and cowboys. 
In our experiment, we evaluate our model on MC500 since there are only 280 questions in the MC160, which is not suitable in our setting.

Appendix~\ref{sec:details} shows more details about both datasets.

\subsection{Model Description}
In this work, we used BERT-base~\citep{devlin2019bert} as
the pre-trained model for both the EQA system and the MCQA system in the following experiments.

 \paragraph{EQA model} The hyperparameters we used are the same as the official released for training SQuAD 1.1. For both datasets, the EQA model is trained on SQuAD 1.1.
\paragraph{MCQA Model} To fine-tune the BERT model on the MCQA datasets, we construct four input sequences, each containing the concatenation of the passage, the question, and one of the choices~\citep{zellers2018swag}. The separator tokens {\tt[SEP]} are added between the passage and the question. Next, we fed the {\tt[CLS]} token representation to the classifier and got the scores for each choice.

%\section{Experiment Results \& Analysis}

\section{Experiment Results}
Table~\ref{tab:results} shows the results of baselines and our methods on RACE and MC500.

\paragraph{RACE} Our methods outperform SW and EQA Match across all the datasets with absolute gain 4$\sim$9\% accuracy, which shows the MCQA model can improve itself from the noisy candidate sets. 
MML and Hard-EM outperform Highest-Only in all cases, which indicates that relying only on the single choice with the highest score is insufficient.
The improvement with EQA Matching Algorithm is more significant than with SW Matching Algorithm. 
This implies Candidates Choosing stage plays a significant role in the performance; more details will be discussed later.

\paragraph{MC500} 
With the SW Matching algorithm, our methods outperform the performance baseline across all the datasets with absolute gains of 1$\sim$5\% accuracy. 
With the EQA Matching Algorithm, because on MC500, EQA has achieved a comparable result with supervised learning, the proposed approaches do not further improve EQA.
The performance of our method drops in MC500-One because EQA models can better capture the information within a sentence than multiple sentences, leading MC500-One performance much better than MC500-Multi with EQA models.
On the other hand, we improve the performance of MC500-Multiple by about 12\%. 
This shows that our method can further improve EQA in the more difficult examples that the EQA model cannot answer correctly. 

%This also shows that our method is not just picking the choice which has the most overlapped between context and the choice. 
%Noticed that we achieved a comparable result to supervised learning on MC500 test-set. 

%In MC500, the results in dev-set and test-set are not as consistent as in RACE. 
%We hypothesize that this is due to the bias in data split.
%There is also a performance gap between dev-set and test-set in supervised learning, which echos the existence of bias. 

\section{Analysis}

\paragraph{Candidate Set \& Matching Methods}
Table~\ref{tab:candidates} shows the average size of candidate sets chosen by EQA and SW Matching, and their \textit{Percent Including Answer}, that is, the percent of candidate set including the correct answer.
The \textit{Percent Including Answer} is much better for SW than EQA on RACE because the candidate sets selected by SW are larger than EQA.
We find that EQA gives more concentrated confidence scores to the choices than SW, leading to smaller candidate sets.  
Although the \textit{Percent Including Answer} of SW is larger than by EQA (Table~\ref{tab:candidates}), the candidates picked by EQA have higher quality than candidates picked by SW, as shown in Table~\ref{tab:chosen}. 

Table~\ref{tab:candidates} implies that MCQA models from the proposed learning strategy do not just randomly choose a prediction from the candidates.
The performance of the proposed approaches in Table~\ref{tab:results} is much higher than the performance of randomly sampling from the candidate set, that is, (B) / (A) in Table~\ref{tab:candidates}. 

\begin{figure}[t!]
    \centering
    \begin{subfigure}[b]{0.45\columnwidth}

    \includegraphics[width=\linewidth]{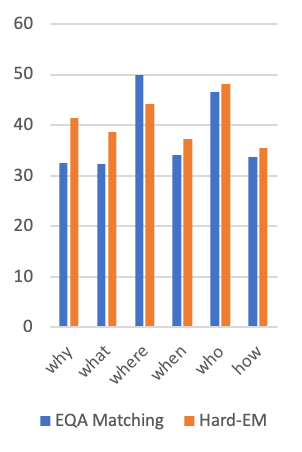}
    \subcaption{EQA Matching and hard-em approach}
    \end{subfigure}
    \begin{subfigure}[b]{0.45\columnwidth}
    \includegraphics[width=\linewidth]{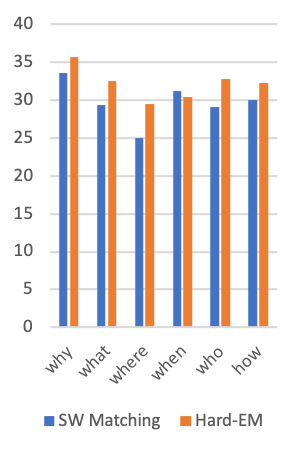}
    \subcaption{SW Matching and hard-em approach}
    \end{subfigure}
    \caption{Accuracy (\%) on different type of question}
    \label{fig:type}
    \vspace{-0.5cm}
\end{figure}

\paragraph{Question Types} 
To see how our learning method works with respect to the type of question, we divided the questions in RACE into six types: why, what, where, when, who, and how. 
We choose to analyze RACE because it has more questions than MC 500.
Figure~\ref{fig:type} shows the accuracy of each question types.
The results show that the proposed approach does not favor specific types of questions. 
We found that no matter the candidate set selection methods, the proposed method improved all types of questions, except "where" for EQA and "when" for SW.
Understanding why some question types do not been improved by unsupervised MCQA in some cases is our future work. 

%The most accuracy improvement comes from "what" questions because it has the largest portion of all. 

%% file: sections/tables/results.tex
\begin{table*}[hbt!]
    \vspace{-0.75cm}
    \centering 
    \small
    % \footnotesize
    \begin{tabulary}{\textwidth}{lcccccc|cccccc}
     \toprule
        & \multicolumn{2}{c}{RACE} 
        & \multicolumn{2}{c}{RACE-M}
        & \multicolumn{2}{c|}{RACE-H}
        & \multicolumn{2}{c}{MC500} 
        & \multicolumn{2}{c}{MC500-One}
        & \multicolumn{2}{c}{MC500-Multi.}
        \\
        & dev & test & dev & test & dev & test & dev & test & dev & test & dev & test\\
     \midrule
     \multicolumn{5}{l}{\textit{Starting from SW Matching Algorithm}}\\
     \midrule
        SW  & 30.8 & 30.2 & 36.2 & 35.2 & 28.4 & 28.1 
            & 46.5 & 42.8 & 36.7 & 43.7 & 54.5 & 42,1 \\
        Highest-Only & 31.8 & 30.8  & 37.5 & 36.4 & 29.4 & 28.5 
                      & 46.0 & 42.3 & 44.4 & 41.5 & 47.2 & 43.0 \\
        MML  & 34.0 & 33.1 & 40.3 & 40.5 & 31.4 & 30.1
             & 50.0 & 45.3 & \bf 46.6 & 44.4 & 52.7 & \bf 46.1\\
        Hard-EM  & \bf 34.3 & \bf 34.0 & \bf 41.0 & \bf 41.2 & \bf 31.5 & \bf 31.0 
                 & \bf 51.5 & \bf 45.7 & 44.4 & \bf 47.7 & \bf 57.3 &  44.0 \\ 
     \midrule
     \multicolumn{5}{l}{\textit{Starting from EQA Matching Algorithm}}\\
     \midrule     
        EQA Match  & 32.3 & 32.2 & 40.3 & 40.5 & 28.9 & 28,8 
            & 62.5 & \bf 64.1 & \bf 75.6 & \bf 80.9 & 51.8 & 49.8\\
        Highest-Only   & 37.0 &	36.9 & 48.8 & 46.1 & 32.1 & 33.1
                      & \bf 67.5 & 60.6 & 67.7 & 66.0 & \bf 67.2 & 56.0\\
        MML  & 38.6 & \bf 39.4 & \bf 49.7 & 49.6 & 34.0 & \bf 35.2 
             & 65.5 &  61.3 &  67.8 & 67.1 & 63.6 & 56.3\\
        Hard-EM  & \bf 39.1 & 39.2 & 49.0 & \bf 49.7 & \bf 35.0 & 34.9 
                 & 66.0 & 63.3 & 68.9 & 66.0 & 63.6 & \bf 60.9 \\
      \midrule
      \midrule
        Supevised & 64.9 & 65.5 & 70.0 & 71.0 & 64.0 & 63.3 
                  & 70.0 & 64.3 & 75.6 & 69.0 & 60.4 & 65.4 \\
     \bottomrule
\end{tabulary}
\caption{\textbf{Results on RACE and MC500 of MCTest.} The evaluation measure is accuracy (\%). The Supervised Learning was training with ground truth and used the same hyperparamter as others.} 
\label{tab:results}
\vspace{-0.25cm}
\end{table*}

%% file: sections/tables/candidates.tex
% \begin{table*}[hbt!]
%     \centering 
%     \small
%     % \footnotesize
%     \begin{tabularx}{\textwidth}{lcccccc|cccccc}
%      \toprule
%         & \multicolumn{2}{c}{RACE} 
%         & \multicolumn{2}{c}{RACE-M}
%         & \multicolumn{2}{c|}{RACE-H}
%         & \multicolumn{2}{c}{MC500} 
%         & \multicolumn{2}{c}{MC500-One}
%         & \multicolumn{2}{c}{MC500-Multi.}
%         \\
%         & dev & test & dev & test & dev & test & dev & test & dev & test & dev & test\\
%      \midrule
%      \multicolumn{5}{l}{\textit{SW Matching Algorithm}}\\
%      \midrule
%         Avg. numbers of candidates  & 32.3 & 32.2 & 40.3 & 40.5 & 28.9 & 28,8 
%                  & 62.5 & \bf 64.1 & \bf 75.6 & \bf 80.9 & 51.8 & 49.8\\
%         Include Answer & 32.3 & 32.2 & 40.3 & 40.5 & 28.9 & 28,8 
%                  & 62.5 & \bf 64.1 & \bf 75.6 & \bf 80.9 & 51.8 & 49.8\\
%      \midrule
%      \multicolumn{5}{l}{\textit{EQA Matching Algorithm}}\\
%      \midrule     
%         Avg. numbers of candidates  & 32.3 & 32.2 & 40.3 & 40.5 & 28.9 & 28,8 
%                  & 62.5 & \bf 64.1 & \bf 75.6 & \bf 80.9 & 51.8 & 49.8\\
%         Include Answer & 32.3 & 32.2 & 40.3 & 40.5 & 28.9 & 28,8 
%                  & 62.5 & \bf 64.1 & \bf 75.6 & \bf 80.9 & 51.8 & 49.8\\
%      \bottomrule
% \end{tabularx}
% \caption{Candidate Set Analysis of RACE and MC500 of MCTest.} 
% \label{tab:candidates}
% \end{table*}

\begin{table}[hbt!]
    \centering 
    % \small
    \footnotesize
    \begin{tabular}{lcccc}
    
     \toprule
        & \multicolumn{2}{c}{RACE} 
        & \multicolumn{2}{c}{MC500} 
        \\
        & dev & test & dev & test \\
     \midrule
     \multicolumn{3}{l}{\textit{SW Matching Algorithm}}\\
     \midrule
        (A) Avg. num. of candidates  & 3 & 3 & 1.98 & 1.85 \\
        (B) Percent Including Ans. & 79.2 & 79.0 & 67.0 & 62.1 \\
        (B) / (A) & 26.4 & 26.3 & 33.8 & 33.6  \\
     \midrule
     \multicolumn{3}{l}{\textit{EQA Matching Algorithm}}\\
     \midrule
        (A) Avg. num. of candidates  & 1.35 & 1.38 & 1.63 & 1.62   \\
        (B) Percent Including Ans. & 40.9 & 41.8 & 73.0 & 71.5  \\
        (B) / (A) & 30.3 & 30.3 & 44.8 & 44.1  \\
     \bottomrule
    \end{tabular}
    \caption{\textbf{The average size of candidate sets chosen by EQA and SW Matching.} \textit{Percent Including Answer} means the percent of candidate set including the labeled answer. (B) / (A) is the accuracy of randomly selecting a choice from a candidate set.}
\label{tab:candidates}
\end{table}

%% file: sections/tables/chosen.tex
\begin{table}[ht!]
    \centering 
    % \footnotesize
    \begin{tabular}{| l|l | cc |}
     \hline
        EQA & SW & RACE-train & MC500-train\\
     \hline
        \ding{51}& \ding{55}  & 29759 & 202 \\
        \ding{55}& \ding{51}  & 8461 & 194 \\
     \hline
    \end{tabular}
\caption{\textbf{Candidate Set Analysis of RACE and MC500 of MCTest.} \textit{Case1: candidates chosen by EQA including the answer but candidates chosen by SW not including the answer.
Case2: candidates chosen by SW including the answer but candidates chosen by EQA not including the answer.}} 
\label{tab:chosen}
\vspace{-0.5cm}
\end{table}

%% file: sections/conclusion.tex
\section{Conclusion}
In this paper, we proposed an unsupervised MCQA method, which exploits the pseudo labels generated by some basic rules or external non-MCQA datasets. 
The proposed method significantly outperforms the baseline approaches on RACE and is even comparable with the supervised learning performance on MC500.
We hope this paper sheds light on unsupervised learning in NLP tasks.

%% file: sections/appendix.tex
\section{Dataset Details}
\label{sec:details}

\begin{table}[hbt!]
    \centering 
    % \footnotesize
    \begin{tabular}{| l | ccc |}
     \hline
        & & RACE & \\
        & train & dev & test \\
        
     \hline
        RACE-M  & 25421 & 1436 & 1436 \\
        RACE-H  & 62445 & 3451 & 3698 \\
     \hline
     \hline
        & & MC500 & \\
        & train & dev & test \\
     \hline
        MC500-One  & 564 & 90 & 277 \\
        MC500-Multi  & 636 & 119 & 323 \\
     \hline
    \end{tabular}
\caption{\textbf{Number of examples in RACE and MC500 of MCTest.}  RACE-M and MC500-One are easier than RACE-H and MC500-Multi separately.} 
\label{tab:details}
\end{table}
\vspace{-0.5cm}
\section{Matching Algorithms}
\label{sec:alg}
\input{sections/alg/sw}

\input{sections/alg/eqa}

\section{Training Details}
We finetuned all models with a linear learning rate decay schedule with 1000 warm-up steps. The batch size is 32, and the max length of the input size is 320. 
For RACE, we set the threshold to 0, the max number of candidates to 3 with SW Matching, and set the threshold to 50, the max number of candidates to 3 with the EQA Matching. For MC500, we set the threshold to 3, the max number of candidates to 2 with SW Matching, and the threshold to 50, the max number of candidates to 3 with the EQA Matching.

Following ~\citet{min2019discrete}, when we use hard-EM as objective, we perform annealing: at
training step $t$, the model use MML as objective with a probability of $\min(t/\tau, 0.8)$ and otherwise use hard-EM, where $\tau$ is a hyperparameter. We tried $\tau=$ 1000, 4000, and 8000.

%% file: sections/alg/sw.tex
\begin{algorithm}[h!]

\caption{Sliding Window}
\label{alg:sw}
\SetKwInOut{Input}{Input}
\SetKwInOut{Define}{Define}
\Input{Threshold $t$, max numbers of candidates $k$, a set of passage words $P$, set of words in question $Q$, and a set of words in choices $C_{1\ldots n}$.}
\Define{$Count(w) \coloneqq \sum_i\mathbbm{1}(P_i=w)$ where $P_i$ is the i-th word in passage $P$;}
\Define{$IC(w) \coloneqq \log\left(1+\frac{1}{Count\left(w\right)}\right)$}

candidates $\gets$ Array[]\\
\For{$i=1$ to $n$}{
  $S \gets C_i \cup Q$
  \[
  \begin{split}
    &\text{score}_i \gets\\
    & \max\limits_{j=1\ldots \lvert P \rvert}  \sum\limits_{w=1\ldots \lvert S \rvert} \begin{cases} IC(P_{j+w}), & \text{if } P_{j+w}\in S \\0, & \text{otherwise} \end{cases} 
  \end{split}
  \]
  \If{$\text{score}_i \geq t$}{
    candidates.append((i, $\text{score}_i$))
  }
}
sort candidates descending by score\\
\Return first $k$ elements of candidates
\end{algorithm}

%% file: sections/alg/eqa.tex
\begin{algorithm}[!ht]
\caption{EQA Matching}
\label{alg:eqa}
\SetKwInOut{Input}{Input}
\SetKwInOut{Define}{Define}
\Input{Threshold $t$, max numbers of candidates $k$, a set of passage words $P$, set of words in question $Q$, and a set of words in choices $C_{1\ldots n}$ and a pre-trained EQA model $M$}

candidates $\gets$ Array[]\\
$A \gets M.predict(P,Q)$\\
\For{$i=1$ to $n$}{
\[
  \text{score}_i \gets \text{Gestalt Pattern Matching}(A, C_i)
\]
  \If{$\text{score}_i \geq t$}{
    candidates.append((i, $\text{score}_i$))
  }
}
sort candidates descending by score\\
\Return first $k$ elements of candidates
\end{algorithm}